\documentclass{article}
\usepackage[utf8]{inputenc}
\usepackage[numbers]{natbib} 
\usepackage{amsmath} 
\usepackage{amsfonts}
\usepackage{amsthm}
\usepackage{bm}
\usepackage{caption}
\usepackage{subcaption}
\usepackage{booktabs}
\usepackage{longtable,tabu}
\usepackage{listings}
\usepackage{geometry}
\usepackage{color} 
\usepackage{graphicx}
\usepackage{tikz}
\usepackage{pgfplots}
\usepackage{hyperref}
\pgfplotsset{compat=1.15}
\usepgfplotslibrary{fillbetween}
\usepgfplotslibrary{colorbrewer}
\usepgfplotslibrary{groupplots}
\usepackage{authblk}

\newtheorem{lemma}{Lemma}
\newtheorem*{lemma-nonum}{Lemma}

\pgfplotsset{
    discard if not/.style 2 args={
        x filter/.code={
            \edef\tempa{\thisrow{#1}}
            \edef\tempb{#2}
            \ifx\tempa\tempb
            \else
                
            \fi
        }
    }
}
\title{Translational Equivariance in Kernelizable Attention}
\date{\vspace{-5ex}}

\begin{document}
\author[$\ast$, 1]{Max Horn}
\author[$\ast$, 1]{Kumar Shridhar}
\author[$\ast$, 1]{Elrich Groenewald}
\author[$\ast$, 1]{Philipp F. M. Baumann}
\affil[$\ast$]{Equal contribution.}
\affil[1]{ETH Zürich, Switzerland}

\maketitle

\begin{abstract}
While Transformer architectures have show remarkable success, they are bound to the computation of all pairwise interactions of input element and thus suffer from limited scalability.
Recent work has been successful by avoiding the computation of the complete attention matrix, yet leads to problems down the line.
The absence of an explicit attention matrix makes the inclusion of inductive biases relying on relative interactions between elements more challenging.
An extremely powerful inductive bias is translational equivariance, which has been conjectured to be responsible for much of the success of Convolutional Neural Networks on image recognition tasks.
In this work we show how translational equivariance can be implemented in efficient Transformers based on kernelizable attention -- Performers.
Our experiments highlight that the devised approach significantly improves robustness of Performers to shifts of input images compared to their naive application.
This represents an important step on the path of replacing Convolutional Neural Networks with more expressive Transformer architectures and will help to improve sample efficiency and robustness in this realm.
\end{abstract}

\section{Introduction}\label{sec:Intro}
Self-attention architectures like Transformers \cite{vaswani.2017} have achieved great success for Natural Language Processing tasks, and the trend seems to continue for computer vision domain \cite{dosovitskiy2020image}. However, a lot of these successes is owed to the ever growing network size \cite{brown2020language}, making them inefficient for training and deployment. The bottleneck lies in the quadratic complexity of the self-attention mechanism, leading to a lot of recent attempts towards making it linear \citep{choromanski.2020, wang2020linformer, kitaev2020reformer, roy2020efficient}. 
Apart from large computation resources needed to train these huge models, gathering labelled data points is an equally serious challenge.
Properties like translational equivariance in  Convolutional  neural  nets  (CNNs) have been found to be a useful design choice that can improve the generalization capabilities of networks on datasets of limited size \cite{mohamed2020data}. In our work, we introduce Relative Positional Encodings (RPE) into the kernel-approximation based method of \citet{choromanski.2020} to induce translational equivariance properties in this efficient transformer variant. We evaluate our methods' performance on several image recognition tasks (\textsc{MNIST}, \textsc{Fashion-MNIST} \citep{xiao2017/online} and \textsc{CIFAR-10} \citep{krizhevsky2014cifar}) and compare our strategies against common CNN architectures (AlexNet \cite{krizhevsky2017imagenet}, VGG \cite{simonyan2014very} and ResNet \cite{he2016deep}) alongside other efficient transformers approaches (Linformer \cite{wang2020linformer} and Routing Transformer \cite{roy2020efficient}).  Further, we show how the introduced translation equivariance properties are beneficial in improving robustness and generalization to image translations.

\section{Models and Methods}\label{sec:MandM}

\subsection{Efficient Transformers} \label{sec:eff-transformers}

The original Transformer \citep{vaswani.2017} introduced, among other results, the multi-headed self-attention
mechanism. This mechanism, for a single head, is characterized by the following computation

\begin{equation}\label{eq:RegAttention}
    Att(\bm{Q},\bm{K},\bm{V}) = \bm{A}\bm{V}, \quad \bm{A} = \text{softmax}(\bm{Q}\bm{K}^\top/\sqrt{d})
\end{equation}

where $\bm{Q} = \bm{x}\bm{W_q}\in \mathbb{R}^{L \times d}$, $\bm{K} = \bm{x}\bm{W_k}\in \mathbb{R}^{L \times d}$, and $\bm{V} = \bm{x}\bm{W_v}\in \mathbb{R}^{L \times d}$, and $\bm{W_q},\bm{W_k}$ and $\bm{W_v}$ are (learnable) weights that project the input
embedding which is equipped with an absolute positional encoding, termed $\bm{x}$, to a dimension $d$. At this point, computing the necessary
attention matrix $\bm{A}$ requires computing $\bm{Q}\bm{K}^\top$ whose time and memory complexity is quadratic in the
input sequence length $L$. Most of the recent literature has addressed this scalability issue by inducing sparsity into
the computation of $\bm{A}$ and thus approximates $\bm{A}$ through different procedures. \citet{tay.2020} provide an
overview over these procedures and divide efficient transformers according to their core technique into efficient
transformers based on fixed/factorized/random patterns, memory, recurrence, learnable patterns and low rank/kernel
approximations. For our project, we focus on kernel approximations, which rewrite the attention mechanism through
kernelization and thus circumvent the explicit computation of $\bm{A}$ in various ways.\\

The kernel approximation technique in the Performer \citep{choromanski.2020} responds to the scalability issue of regular
Transformers by introducing the FAVOR algorithm to estimate the regular full-rank attention. While previous approaches
relied on structural assumptions for the attention matrix such as sparsity and low-rankness or did not approximate the original softmax function \cite{katharopoulos2020transformers}, the FAVOR method approximates the
attention matrix by using random feature maps which in expectation converge to the softmax operation. In general, FAVOR’s random feature
map is a vector with dimension $m’ = m \times l$ which contains the mapped scalars of $j = 1,\ldots, l$ real functions
where each $f_j$ takes the dot-product between $m$ draws from a multivariate standard Gaussian distribution and either
the rows $q_i$ of $\bm{Q}$ or $k_i$ of $\bm{K}$ as an input, for each element $i = 1,\ldots,L$ of the input sequence. This vector is then scaled by $\frac{h(q_i)}{\sqrt{m}} or
\frac{h(k_i)}{\sqrt{m}}$ respectively, where $h$ is some deterministic function of $q_i$ or $k_i$. By means of $\phi$,
\citep{choromanski.2020} show that one can rewrite some non-linear kernel function $K$, in our case the exponential/softmax kernel, by
the expectation (i.e.  a linear operator) of a dot-product between $\phi(q_i)$ and $\phi(k_i)$ respectively. This works
for various kernel-functions, however, for the softmax-kernel it is required to set $h(x)= exp(\left\Vert
x \right\Vert^2 / 2)$, $f_1 = sin$ and $f_2 = cos$ (i.e. two positional features) where $x$ is either $q_i$ or $k_i$ so
that we are able to obtain the corresponding $\phi$. Afterwards, we can compute $\phi(q_i) = q_i’$ and $\phi(k_i)
= k_i’$ for each $i$ independently such that we are able to obtain $\bm{K}’$ and $\bm{Q}’$. As a result, the regular Attention
$exp(\bm{Q}\bm{K}^\top)$ can be approximated by $\bm{Q}’\bm{K}’^{\top}$. This is the crucial part to achieve linear complexity in $L$
(assuming $d < L$) since FLAVOR enables to compute $\bm{K}’^{\top} \bm{V}$ before it is multiplied with $\bm{Q}’$. As a result, it
circumvents the costly computation of $\bm{Q}\bm{K}^{\top}$ of \eqref{eq:RegAttention} and, further, allows a decomposition that
provides an unbiased estimator for the Attention matrix that turns more precise, the more random features $m$ are
sampled.

\subsection{Translational equivariance}
Following the notation of \citet{gordon2020conv}, a function $\Phi: \mathcal{Z} \rightarrow \mathcal{Z}$ is translation
equivariant to a translation operation $T_\tau: \mathcal{X} \times \mathcal{Z} \rightarrow \mathcal{Z}$ if $\Phi(T_\tau
Z) = T_\tau \Phi(Z)$ for all $\tau \in \mathcal{X}$ and $Z \in \mathcal{Z}$.  For example, assume $Z$ to be
a collection of $L$ measurements $y$ at locations $x$, such that $Z = ((x_1, y_1), \ldots, (x_L, y_L))$ and define the
translation operation as $T_\tau Z = ((x_1 - \tau, y_1), \ldots, (x_L - \tau, y_L))$, then $\Phi$ would be translation
equivariant with respect to shifts of measurement locations if first shifting the measurements and then applying the
function (i.e. $\Phi(T_\tau Z)$) would lead to the same result as first applying the function an then shifting (i.e.
$T_\tau \Phi(Z)$).

In machine learning, equivariances play an important role in ensuring generalization to domains of know structure.
A model is often referred to having a ``good inductive bias'' if properties and structures of the model are in
line with a certain type of task or data domain \cite{cohen16group}.
Translation equivariance can thus be seen as an inductive bias which encompasses that the learnt function should be
equivariant to translations of its input data.  This inductive bias has partially attributed to the famous success of
CNNs on image classification tasks \citep{lecun1998gradient,cohen16group, mohamed2020data}.

\subsection{Translational Equivariant Performers}
Since Transformers with self-attention are completely permutation equivariant \cite{lee2019set} they cannot gain
positional information from a location of an element in the data structure such as CNNs.
Transformers instead usually rely on the absolute positional embeddings, which encode absolute position of an element into
a distributed representation (usually via sinusoidal embeddings) and are added to the input embedding prior to the
application of the first self-attention layer.  Encoding the absolute position into the representation prevents the
learnt function from exhibiting beneficial properties such as translation equivariance.
In contrast, \citet{dai.2019} rely on relative positional embeddings in order to ensure the transfer of models trained on
shorter sequences to longer sequences an approach which would allow the construction of translation equivariant
transformer architectures. Unfortunately, their approach cannot be easily applied to the most recently suggested
efficient transformer architectures.

In the following, we demonstrate two ways on how to incorporate RPE while
retaining the requirements of the Performer's \cite{choromanski.2020} attention mechanism. We stick to the notation of
\cite{dai.2019, choromanski.2020} and start with the computation of the pre-attention matrix.  If absolute positional encodings $u$ are added to the input embeddings $e$, such that $x_i = e_i + u_i$, the pre-attention matrix computation can be described with
\begin{align}
\tilde{\mathbf{A}}_{i, j}
    &= \langle(e_i + u_i)\bm{W}_q, (e_j + u_j)\bm{W}_k \rangle 
     = (e_i + u_i)\bm{W}_q\bm{W}_k^\top(e_j + u_j)^\top \nonumber \\
    &= e_i\bm{W}_q\bm{W}_k^\top e_j^\top + e_i\bm{W}_q\bm{W}_k^\top u_j^\top + u_i \bm{W}_q \bm{W}_k^\top e_j^\top + u_i \bm{W}_q\bm{W}_k^\top u_j^\top \label{eq:expanded-attn}
\end{align}
%


\paragraph{Strategy 1}
In our first strategy, we discard interaction terms between the content embeddings $e$ and the absolute positions $u$ in Equation~\ref{eq:expanded-attn} and follow the notion, that equivariance is linked to parameter sharing
\cite{ravanbakhsh2017equivariance}.  In particular, we examine how absolute positional encodings can be adapted in an
attention mechanism, such that the resulting pre-attention (i.e. $u_i \bm{W}_q\bm{W}_k^\top u_j^\top$) is independent of the absolute position and solely depends on
the relative positions of the two elements under consideration.  Such a construction will be sufficient to make the
whole transformer architecture translation equivariant.
In the following, we show that given certain constraints, the inner product of projected absolute positional encodings
is invariant to their absolute position and only depends on their relative positions.

%

\begin{lemma}
    Given absolute positional encodings $\phi(z) = [\sin(\omega z), \cos(\omega z)]^\top$ and two projection matrices
    matrices $\bm{W}_q \in \mathbb{R}^{2 \times 2}$ and $\bm{W}_k \in \mathbb{R}^{2 \times 2}$ it holds that $\langle \phi(z
    - \tau)\bm{W}_q, \phi(z' - \tau) \bm{W}_k \rangle = \langle \phi(z) \bm{W}_q, \phi(z') \bm{W}_k \rangle$ for $z, z' \in \mathcal{Z}$ if
    $\bm{W}_q^\top \bm{W}_k$ is of the following form
\begin{equation}
    \bm{W}_q \bm{W}_k^\top = \begin{bmatrix}
        \alpha & \beta\\
        -\beta & \alpha
    \end{bmatrix}
\end{equation}
    where $\alpha, \beta \in \mathbb{R}$.
\end{lemma} \label{lemma:rel-constraint}

\begin{proof}
    The proof can be found in the supplementary materials.
\end{proof}

In our experiments we explore the simplest scenario where $\bm{W}^*_q = \begin{bmatrix} \alpha & \beta\\ -\beta & \alpha
\end{bmatrix}$ and $\bm{W}^*_k = \mathbb{I}_2$ is set to be the identity matrix.  Further, we learn multiple length scales
$\omega_1, \ldots, \omega_m$ for the computation of positional encodings and apply the same restriction to each length
scale individually.
Assuming the features are concatenated to a single vector $\phi(x) = \left[\sin(\omega_1 x), \cos(\omega_1 x), \ldots, sin(\omega_m x), cos(\omega_m
x)\right]$, this can be implemented via a block-diagonal matrix of the form
\begin{equation*}
    \bm{W}^*_q = \text{blockdiag}\left(
    \begin{bmatrix}
        \alpha_1 & \beta_1 \\
        -\beta_1 & \alpha_1
    \end{bmatrix},
    \ldots,
    \begin{bmatrix}
        \alpha_m & \beta_m \\
        -\beta_m & \alpha_m
    \end{bmatrix}
    \right)
\end{equation*}
and $\bm{W}^*_k = \mathbb{I}_{2m}$. Finally, can entirely embed these constraints into the Performer architecture by
augmenting the features prior to the approximation of the softmax operation via random features.  This is achieved by
simply concatenating features derived from the content embedding ($\bm{Q}_c = \bm{e}\bm{W}_q$ and $\bm{K}_c = \bm{e} \bm{W}_k$) and the features from the positional
embedding ($\bm{Q}_p = \bm{u} \bm{W}^*_q$ and $\bm{K}_p = \bm{u} \bm{W}^*_k$) such that $\bm{Q} = \text{concat}(\bm{Q}_c, \bm{Q}_p)$ and $\bm{K} = \text{concat}(\bm{K}_c, \bm{K}_p)$.  We do not
include any positional information in the values $V$.
It is important to note though, that while our construction is invariant to the absolute position in expectation, this
might entirely not the case for any set of random features.  Nevertheless, we conjecture that due to redrawing of the
random features during training the above derived properties should still hold.
Finally, this construction does lead to the same overall runtime complexity of $\mathcal{O}(N)$ as the original
Performer model, but does not allow querying (relative) positional information conditional on the content of a pixel.

\paragraph{Strategy 2}
Due to strategy 1's limitation of having no content-position interactions, for our strategy 2 we take inspiration from \citet{shaw2018self}, in which each element of the pre-attention matrix $\tilde{\mathbf{A}}$ is defined as 
\begin{equation}
\mathbf{\tilde{A}}_{i j}=e_i \mathbf{W}_{q} ( e_j \mathbf{W}_{k} + \mathbf{a}_{i j})^\top=\underbrace{e_i \mathbf{W}_{q} \mathbf{W}_{k}^\top e_j^\top }_{(a)}+\underbrace{e_i \mathbf{W}_{q} \mathbf{a}^\top_{i j}}_{(b)}
\label{eq:shaw}
\end{equation}
Here, $\mathbf{a}_{i j}$ is the learned relative positional encoding for the relative distance between positions $i$ and $j$. These learned relative positional encodings are clipped to a maximum relative distance $k$. i.e. 
\begin{align*}
\mathbf{a}_{i j} &=\mathbf{w}_{\operatorname{clip}(j-i, k)} \\
\operatorname{clip}(x, k) &=\max (-k, \min (k, x))
\end{align*}

We use the relative pixel distance between pixel $i$ and pixel $j$ as the relative distance between input positions $i$ and $j$. This is calculated as the absolute difference in $x$ position plus the absolute difference in $y$ position of the two pixels. We clip this relative pixel distance to a maximum distance $k$ and learn the relative position encodings $\mathbf{w}_1, \mathbf{w}_2, \ldots, \mathbf{w}_k$.

To combine this approach with the attention approximation in \citet{choromanski.2020}, we split (\ref{eq:shaw}) into two separate attention computations, and assign these to different attention heads in the multi-head self attention (\citep{vaswani.2017}). Half of the attention heads in each self-attention layer use content-based attention defined by $(a)$ and the remaining attention heads use relative position-based attention defined by $(b)$. The output of the content-based attention defined by $(a)$ can be approximated directly as in \citet{choromanski.2020}, while output for relative position-based attention is approximated as follows.

Let $\mathbf{q}_i^{\prime} := \phi(\mathbf{q}_{i}^{\top})^{\top}$ and $\mathbf{a}_{ij}^{\prime} := \phi(\mathbf{a}_{ij}^{\top})^{\top}$, where $\phi(\cdot)$ is the random feature map as defined by \citet{choromanski.2020} and $\mathbf{q}_i$ is the $i$th row-vector of the query matrix $Q$. The $i$th row of the attention output for relative position based attention is calculated as
$$
\sum_{j=1}^L \langle\mathbf{q}_i^{\prime}, \mathbf{a}_{ij}^{\prime} \rangle\mathbf{v}_j 
$$

A naive way to compute this is to compute $\langle\mathbf{q}_i^{\prime}, \mathbf{a}_{ij}^{\prime} \rangle$ for each pair $(i, j)$, which would have quadratic cost in the sequence length L. However, for small clipping distance $k$, most $\mathbf{a}_{ij}$ are equal to $\mathbf{w}_k$, so the computation can be done more efficiently by setting 
$$
\sum_{j=1}^L \langle\mathbf{q}_i^{\prime}, \mathbf{a}_{ij}^{\prime} \rangle\mathbf{v}_j 
= \langle\mathbf{q}_i^{\prime} ,\mathbf{w}_k^{\prime}\rangle \sum_{j=1}^L\mathbf{v}_j 
+ \sum_{m} \langle\mathbf{q}_i^{\prime}, \mathbf{a}_{im}^{\prime}-\mathbf{w}_{k}^{\prime} \rangle\mathbf{v}_m 
$$
where the final term sums over all positions $m$ that are within clipping distance $k-1$ from $i$. This allows the attention output to be computed with $O(Lk^2)$ cost instead of $O(L^2)$.

\section{Results}\label{sec:Results}

\begin{table*}[tbp]
\small\centering

\begin{tabular}
    { l  c  c  c  c  c  c  c}
    \toprule
    & \multicolumn{3}{c}{\textsc{Classification (Acc)}} & & & \\
    \cmidrule(lr){2-4}
    \textsc{Model}                     & \textsc{MNIST} & \textsc{FashionMNIST} & \textsc{CIFAR-10} & \textsc{Params} \\
    \midrule
    AlexNet                   & 99.07 &   89.83      &   67.99  & 2.5M  \\
    VGG-16                    & \textbf{99.34} &   \textbf{91.98}      &   \textbf{75.45}  & 134M  \\
    ResNet-18                 & \textit{99.18} &   90.55      &   \textit{70.70}  & 11.2M \\
    \cmidrule{1-1}
    Linformer                 & 97.80 &   89.51      &   53.80  & 10.2M \\
    Routing Transformer       & 96.40 &   87.43      &   56.67  & 11.4M \\
    \cmidrule{1-1}
    Performer (No Pos.)       & 45.0  &   67.3       &   52.7   &  4.7M \\
    Performer (Absolute)      & 98.8  &   \textit{90.6}       &   69.4   &  4.7M \\
    Performer (Rel. S1)       & 98.8  &   90.4       &   66.4   &  4.7M \\
    Performer (Rel. S2)       & 94.7  &   86.7       &   61.4   &  4.5M \\
\bottomrule
\end{tabular}

    \caption{Performance comparison on the test set of different approaches on three Image classification datasets:
    \textsc{MNIST}, \textsc{Fashion-MNIST} and \textsc{CIFAR-10}.  We compare our suggested approaches to both
    CNN baselines (AlexNet, VGG-16 and ResNet-18) two scalable transformer baselines
    (Linformer, Routing Transformer).  The Performer variants are further characterized by the approach used for
    positional encodings, where \texttt{No Pos.} corresponds to no positional encoding, \texttt{Absolute} to absolute
    positional encodings added to the input embedding before the first transformer layer and \texttt{Rel.\ S1} and
    \texttt{Rel.\ S2} correspond to the RPE strategies 1 and 2 respectively.}
\label{tab:setupCresults}
\end{table*}

We compare the effectiveness of our approach on three image classification dataset, namely \textsc{MNIST},
\textsc{Fashion-MNIST} and \textsc{CIFAR-10}. In order to homogenize the setup between the different datasets and to
allow utilization of similar
architectures between the experiments, we resize the images of \textsc{MNIST} and \textsc{Fashion-MNIST} to $32 \times
32$ pixels prior to the application of our models. For further information we refer the interested readers to section
\ref{sec:dataset-desc} of the appendix. Due to computational constraints, we could not demonstrate the results on
\textsc{ImageNet64}.

To demonstrate the effectiveness of relative positional embedding in Performer style transformer architecture, we
compare our two approaches \texttt{Rel.\ S1} and \texttt{Rel.\ S2} to no positional embedding \texttt{No Pos.} and
absolute positional \texttt{Absolute} version of Performers. For \texttt{Rel.\ S2} we tried clipping distances 2, 4,
and 6. The best results were achieved with clipping distance 6, which are given in our comparison. A detailed list of
hyperparameters can be found in section~\ref{sec:hyperparameters} of the appendix.

\begin{figure}[tbp]
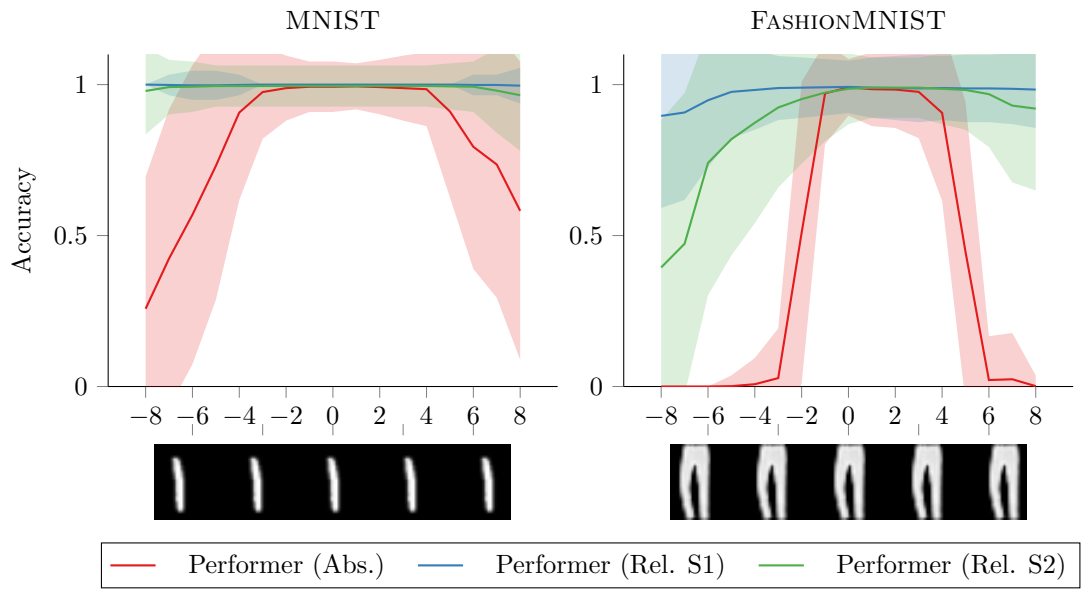

    \include{shift_figure}
    \caption{Performance evaluation on subset of test set of \textsc{MNIST} and \textsc{Fashion-MNIST} under object
    shifts along the x-axis.  For both datasets we solely consider the subset of images which are associated with the classes ``digit
    1'' or ``pants'', respectively and which can be shifted for 8 pixels in both directions without removing non-zero
    pixels. The shaded area corresponds to the standard deviation of the metrics.
    } \label{fig:shift-exp}
\end{figure}

Our approach achieves similar performance ($\pm1$ absolute value) on \textsc{MNIST} and \textsc{Fashion-MNIST} dataset when compared to the CNN architectures and to \texttt{Absolute} positional encoding. Upon comparing with efficient versions of Transformers (Linformer and Routing Transformer), Relative Performer achieves $+1\ \text{to}\ 2$ absolute better performance and the difference goes up to $+30\ \text{to}\ 50$ when compared to \texttt{No Pos.} version. The difference is more evident for \textsc{CIFAR-10} dataset when \texttt{Rel.\ S1} is compared to CNN architectures with CNNs outperforming our approach by $+3\ \text{to}\ 8$ on absolute numbers. However, the \texttt{Absolute} Performer performance falls behind the CNNs ($-1\ \text{to}\ 6$) too and it is a no surprise for our approach to fall behind alongside too. However, Performer architecture uses 2X lesser parameters than ResNet and other efficient Transformer versions: Linformer and Routing Transformer and a massive 30X fewer parameters than VGG-16 network. Performer on the other hand uses twice the number of parameters when compared to AlexNet and the performance is comparable to that (even for \textsc{CIFAR-10}). However, our approach outperforms other Transformer variants: Linformer, Routing Transformer and \texttt{No Pos.} by $+15\ \text{to}\ 20$ despite fewer parameters.

In order to further explore the benefit of translational equivariance for image classification, we evaluated the
trained models on shifted versions of the \textsc{MNIST} digit 1 and the \textsc{Fashion-MNIST} class ``pants'' as shown in
Figure~\ref{fig:shift-exp}.  These shifted versions of the image can be considered slightly out of distribution
compared to the training distribution where the objects are always centered into the middle of the image.  Our
hypothesis was that models with translational equivariance should be considerably more robust to these out
of distribution samples. In Figure~\ref{fig:shift-exp} it is apparent that the Performer relying purely on absolute
positional encodings degrades in performance after shifting the images only a few pixels to the left or to the right on
both datasets.  For the ``pants'' class of the \textsc{Fashion-MNIST} dataset, the accuracy drops to almost $0$ after
shifting only $3$ pixels to the left where as for the \textsc{MNIST} digit 1 a drastic performance decrease only occurs after
shifting more than 5 pixels. In both scenarios the model with translational equivariance is much more robust to these
shifts. Both strategies 1 and 2 show practically no deterioration for \textsc{MNIST} and a much later decrease in performance on
\textsc{Fashion-MNIST}.  Further, strategy 1 is more robust to shifts on \textsc{Fashion-MNIST} (showing approx.\ $80\%$ accuracy after
a shift of 8 pixels to the left) than strategy 2 where accuracy drops to approx.\ $45\%$ for the same perturbation.

\section{Discussion}\label{sec:Disc}

We were able to show that the suggested approach can be considered a competitive alternative to CNN architectures.  Nevertheless, we were not able to reach the performance of CNNs on the \textsc{CIFAR-10}
dataset, indicating the necessity of further research.  One possible explanation is that CNNs additionally enforce
locality at different levels of the feature hierarchy, which is not the case for our approach as it allows information
to be transferred globally.  Further, while both approaches can be considered comparable in terms of upper bound
runtime, the constants of these bounds still differ.  Finally, in practical use cases convolution operations are
implemented via highly optimized and efficient GPU kernels such that CNNs still train orders of magnitude faster than
any of the scalable Transformer approaches presented in this work.

Comparing our two suggested strategies, we see slight performance differences between them.  In particular strategy
2 performs slightly worse than strategy $1$ in all performance assessments.  This could potentially be traced back to
the main difference between the approaches: While S1 allows content-content (pixel value to pixel value) interactions
and position-position interactions, S2 allows content-content and content-position interactions.  It could thus be
argued that approach S1 is more closely related to a CNN architecture, where position-position interactions and
content-position interactions can be implemented.  Finally, the necessity of clipping the maximal relative distance in S2
could also possibly lead to lower performance.

Finally, it is worth noting that our suggested strategies never out-perform and often do not reach the performance of
the Performer with absolute positional embeddings.  Here a possible avenue of explanation could be that the datasets do
not by themselves require translational equivariance as they are homogenized to a large degree (for example all
\textsc{MNIST} and \textsc{Fashion-MNIST} instances are perfectly centered).  Thus the absolute position of a digit
might actually be beneficial for the model in order to reach high performance and the true benefits of these models in
terms of robustness and generalization capabilities would only be highlighted in more diverse settings or when
transferring models between domains.

Our work indicates potential avenues for the fusion of two research areas which have remained independent for a long
time: the introduction of equivariances/inductive biases into machine learning models and scalable Transformer
architectures.  We are confident that this is only the first step to the creation of new models which are
simultaneously scalable and robust.


\section{Summary}\label{sec:Res}
In this work, we presented two strategies to induce translational equivariance properties in Performer architectures while retaining their major advantage - linear scalability. Furthermore, we experimentally demonstrated that the induced translational equivariance properties are robust to image shifts and hence, can be used as an alternative to CNNs.


\addcontentsline{toc}{section}{References}

\clearpage
\bibliographystyle{plainnat}
\bibliography{bibliography}

\clearpage
\appendix
\section{Appendix}
\subsection{Proof of Lemma~\ref{lemma:rel-constraint}} \label{appx:proof}

\begin{lemma-nonum}
    Given absolute positional encodings $\phi(z) = [\sin(\omega z), \cos(\omega z)]^\top$ and two projection matrices
    matrices $W_q \in \mathbb{R}^{2 \times 2}$ and $W_k \in \mathbb{R}^{2 \times 2}$ it holds that $\langle W_q \phi(z
    - \tau), W_k \phi(z' - \tau) \rangle = \langle W_q \phi(z), W_k \phi(z') \rangle$ for $z, z' \in \mathcal{Z}$ iff
    $W_q^\top W_k$ is of the following form
\begin{equation}
    W_q W_k^\top= \begin{bmatrix}
        \alpha & \beta\\
        -\beta & \alpha
    \end{bmatrix}
\end{equation}
    where $\alpha, \beta \in \mathbb{R}$.
\end{lemma-nonum}

\begin{proof}
\begin{align*}
    \langle \phi(z - \tau) W_q , \phi(z' - \tau) W_k \rangle &= \phi(z - \tau) W_q W_k^\top \phi(z' - \tau)^\top \\
    &= [\alpha \sin(\omega(z - \tau)) + \beta \cos(\omega(z - \tau)), \beta \sin(\omega(z - \tau)) + \alpha \cos(\omega(z - \tau))] \phi(z' - \tau) \\
    &= \alpha \sin(\omega(z - \tau)) \sin(\omega(z' - \tau)) + \beta \sin(\omega(z - \tau))\cos(\omega(z' - \tau)) \\
    &\quad-\beta \cos(\omega(z - \tau))\sin(\omega(z' - \tau)) + \alpha \cos(\omega(z - \tau))\cos(\omega(z' - \tau))\\
    &= \frac{\alpha}{2}  \left[ \cos(\omega(z - \tau) - \omega(z' - \tau)) - \cos(\omega(z - \tau) + \omega(z' - \tau))\right] \\
    &\quad + \frac{\beta}{2} \left[ \sin(\omega(z - \tau) + \omega(z' - \tau)) + \sin(\omega(z - \tau) - \omega(z' - \tau)) \right] \\
    &\quad + \frac{-\beta}{2} \left[ \sin(\omega(z - \tau) + \omega(z' - \tau)) - \sin(\omega(z - \tau) - \omega(z' - \tau)) \right] \\
    &\quad + \frac{\alpha}{2} \left[ \cos(\omega(z - \tau) - \omega(z' - \tau)) + \cos(\omega(z - \tau) + \omega(z' - \tau)) \right] \\
    &= \alpha \cos(\omega(z - \tau) - \omega(z' - \tau)) + \beta \sin(\omega(z - \tau) - \omega(z' - \tau) \\
    &= \alpha \cos(\omega z - \omega z') + \beta \sin(\omega z - \omega z')
\end{align*}
\end{proof}
In the third step trigonometric identities for products of trigonometric functions were applied. This shows that the inner product is invariant under shifts of the inputs.
$\alpha \cos(\omega z - \omega z') + \beta \sin(\omega z - \omega z') = \langle W_q \phi(z), W_k \phi(z') \rangle$
follows by applying the steps of the proof in reverse, whilst assuming $\tau = 0$.

\subsection{Model Hyperparameters} \label{sec:hyperparameters}
We trained all Performer models with the hyperparameters learning rate = $0.0005$, batch size = $22$, attention heads
= $8$, depth = $6$, dim = $256$, attention dropout = $0.1$, feed forward dropout = $0.1$. In the case of strategy 1 we
used $4$ lengthscales for the positional embeddings which were initialized in a similar manner as conventional absolute
positional encodings. For strategy 2 we further use clipping distance = $6$.

\subsection{Dataset description} \label{sec:dataset-desc}
\paragraph{MNIST}
The MNIST dataset of handwritten digits consists of 60,000 training and 10,000 validation images of 28 by 28 pixels. Each image is labelled with its corresponding class number (between zero and nine, inclusive).
\paragraph{Fashion MNIST}
This dataset is similar to the MNIST dataset, except instead of digits it has equivalent class from fashion clothing. It consists of a training set of 60,000 examples and a test set of 10,000 examples, similar to MNIST dataset. Each example is a 28x28 grayscale image, associated with a label from 10 classes.
\paragraph{CIFAR-10}
The CIFAR-10 dataset consists of 60,000 colour images in 10 classes, with 6,000 images per class, each image 32 by 32 pixels large. Each of the classes has 5,000 training images and 1,000 validation images. 
\end{document}